\documentclass[10pt,twocolumn,letterpaper]{article}
\pdfoutput=1 
\usepackage{graphicx}
\usepackage[]{iccv}      

\newcommand{\dashedrightarrow}{\tikz[baseline=-0.5ex]\draw[dashed,->](0,0)--(0.4,0);}
\usepackage{multirow}
\usepackage{ulem}
\usepackage{colortbl}
\usepackage{xcolor}

\definecolor{iccvblue}{rgb}{0.21,0.49,0.74}
\usepackage[pagebackref,breaklinks,colorlinks,allcolors=iccvblue]{hyperref}
\usepackage{subcaption}
\usepackage{tikz}
\usepackage{amsthm}

\newtheorem{definition}{Definition}

\def\paperID{} 
\def\confName{ICCV}
\def\confYear{2025}

\title{Treble Counterfactual VLMs: A Causal Approach to Hallucination}

\author{
Shawn Li$^1$, Jiashu Qu$^2$, Yuxiao Zhou$^3$, Yuehan Qin$^1$, Tiankai Yang$^1$, Yue Zhao$^1$\\
$^1$University of Southern California  
\quad $^2$University of Cincinnati \quad $^3$ National University of Singapore\\  
{\tt\small \{li.li02, yuehanqi, tiankaiy, yzhao010\}@usc.edu,} \\
{\tt\small quju@mail.uc.edu,} \\
{\tt\small e1011019@u.nus.edu} 
}

\begin{document}
\maketitle
\begin{abstract}
Vision-Language Models (VLMs) have advanced multi-modal tasks like image captioning, visual question answering, and reasoning. However, they often generate hallucinated outputs inconsistent with the visual context or prompt, limiting reliability in critical applications like autonomous driving and medical imaging. Existing studies link hallucination to statistical biases, language priors, and biased feature learning but lack a structured causal understanding.
In this work, we introduce a causal perspective to analyze and mitigate hallucination in VLMs. We hypothesize that hallucination arises from unintended direct influences of either the vision or text modality, bypassing proper multi-modal fusion. To address this, we construct a causal graph for VLMs and employ counterfactual analysis to estimate the Natural Direct Effect (NDE) of vision, text, and their cross-modal interaction on the output. We systematically identify and mitigate these unintended direct effects to ensure that responses are primarily driven by genuine multi-modal fusion. Our approach consists of three steps: (1) designing structural causal graphs to distinguish correct fusion pathways from spurious modality shortcuts, (2) estimating modality-specific and cross-modal NDE using perturbed image representations, hallucinated text embeddings, and degraded visual inputs, and (3) implementing a test-time intervention module to dynamically adjust the model's dependence on each modality. Experimental results demonstrate that our method significantly reduces hallucination while preserving task performance, providing a robust and interpretable framework for improving VLM reliability. To enhance accessibility and reproducibility, our code is publicly available at 
\url{https://github.com/TREE985/Treble-Counterfactual-VLMs}.
\end{abstract}    
\section{Introduction}
\begin{figure*}[t]
    \centering
    \includegraphics[width=\textwidth]{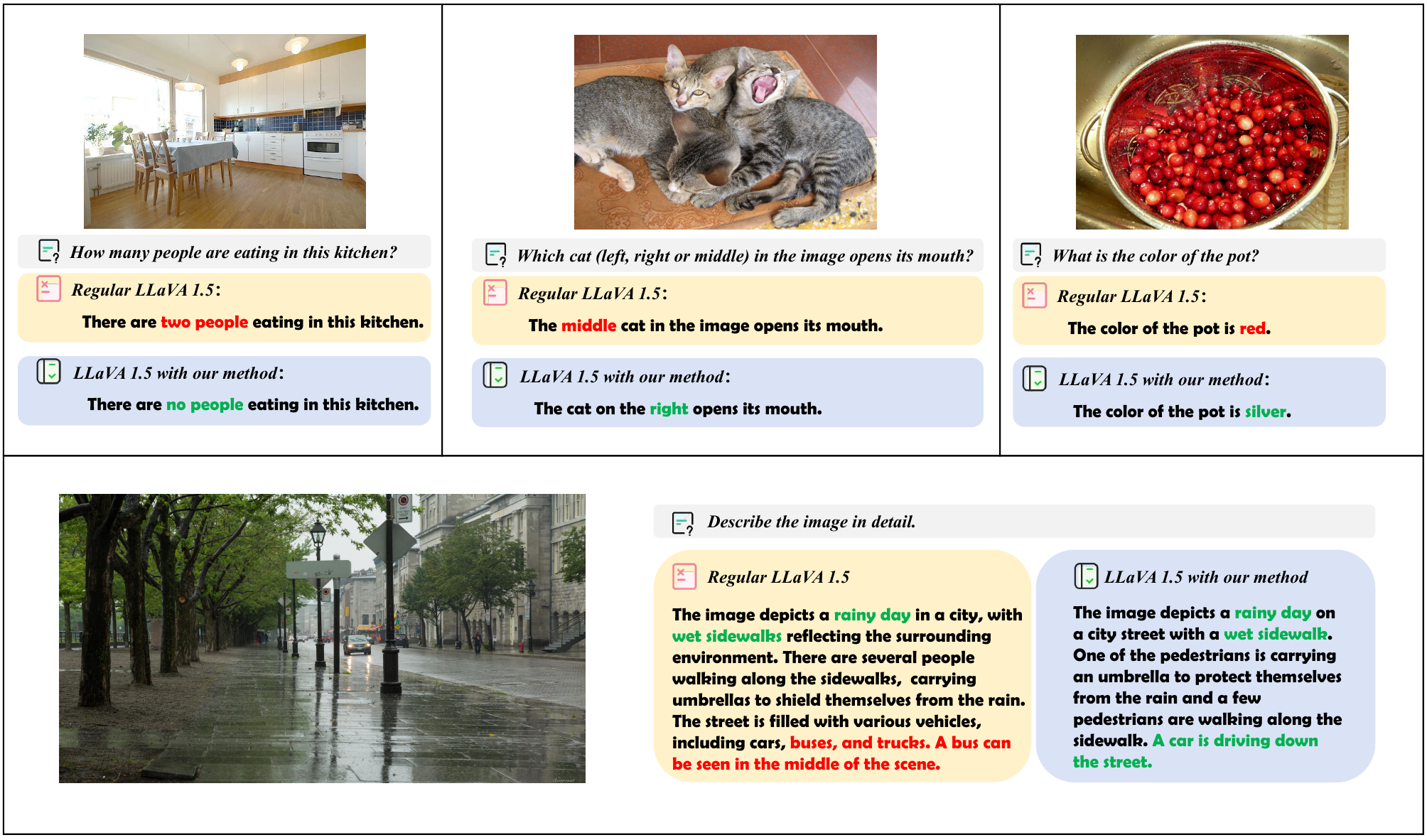}
    \caption{Case study illustrating the impact of our method on VLM hallucination. The figure compares outputs from the original model and our enhanced approach, highlighting reductions in hallucinated content and improved alignment with the visual context. Our method effectively mitigates incorrect  descriptions by refining modality interactions, leading to more accurate and reliable multi-modal reasoning.}
    \label{fig:case_study}
\end{figure*}
Vision-Language Models (VLMs) have achieved remarkable progress in various multi-modal tasks \cite{li2023blip,alayrac2022flamingo}, such as image captioning \cite{mokady2021clipcap}, visual question answering (VQA), and visual reasoning \cite{liu2023visual,radford2021learning}. By integrating visual and textual modalities, VLMs can generate descriptive textual outputs based on image inputs, enhancing machine perception and understanding of multi-modal contexts \cite{chowdhery2023palm}. These models typically consist of a vision encoder, responsible for extracting meaningful features from images, and a language model that processes textual inputs and generates outputs conditioned on both modalities. Recent advancements in large-scale pre-training and transformer-based architectures have significantly improved the generalization ability of VLMs \cite{zhai2022scaling}, making them an essential component in modern AI applications.

Despite their impressive capabilities, VLMs are prone to hallucination \cite{ji2023survey}, a phenomenon where the model generates outputs that are inconsistent with the given visual context or textual prompt. This issue arises when the model produces descriptions, answers, or captions that contain incorrect, fabricated, or misleading information. Hallucination in VLMs can significantly impact their reliability, especially in critical applications such as medical imaging \cite{goddard2023hallucinations}, autonomous driving \cite{chen2024end}, and surveillance \cite{zhao2020reducing}. The underlying causes of hallucination remain a complex and open research question, making it crucial to investigate further and develop effective mitigation strategies.

Several studies have attempted to analyze and mitigate hallucination in VLMs \cite{ji2023survey,zhou2023analyzing,rohrbach2018object,yang2025vidlbeval}, leading to different explanations and strategies. Some works attribute hallucination to statistical biases in pre-training datasets \cite{zhou2023analyzing}, where models memorize spurious correlations rather than genuinely understanding visual-text relationships. Others suggest that VLMs tend to over-rely on language priors \cite{yang2025vidlbeval,rohrbach2018object}, causing them to generate text-based responses that do not sufficiently incorporate visual information. Additionally, some research highlights the role of biased feature learning \cite{kayhan2021hallucination,chen2024multi}, where certain visual or textual patterns dominate the learned representations, leading to distorted multi-modal reasoning. However, these approaches primarily analyze hallucination from a statistical or empirical perspective and often do not differentiate between VLMs and large language models (LLMs), neglecting the unique challenges posed by the multi-modal structure of VLMs.

In this work, we introduce a novel causal perspective \cite{neuberg2003causality} to analyze and mitigate hallucination in VLMs. We construct a causal graph for VLMs, hypothesizing that hallucination arises due to unintended direct influences from either the vision or text modality, bypassing the intended multi-modal fusion process. Specifically, each modality can have independent direct effects on the output, leading to inconsistencies between generated answers and their intended multi-modal context. Based on this causal assumption, we employ counterfactual analysis \cite{lewis2013counterfactuals} to estimate the Natural Direct Effect (NDE) \cite{robins1992identifiability} of each modality. Our goal is to systematically remove each modality’s unintended direct influence from the output, ensuring that the generated response is primarily driven by the joint vision-text fusion process. By reducing these direct effects, we aim to mitigate hallucination in VLMs and improve their overall reliability.

To tackle the hallucination problem in VLMs, we propose a three-step methodology. First, we design structural causal graphs \cite{neuberg2003causality} to model the relationships between vision, text, and generated outputs, distinguishing the correct fusion pathway from spurious modality shortcuts. Second, we introduce a systematic approach to estimate the Natural Direct Effect of vision, text, and their cross-modality interaction. For the vision modality, we generate perturbed images by applying multiple random masks, compute their latent representations, and average them to obtain a perturbed representation. We then compare this perturbed representation against the original to quantify the vision’s NDE. Similarly, for the text modality, we generate hallucinated captions using a language model \cite{zhao2023survey}, extract their representations, and compute the difference between the original and hallucinated text embeddings. To assess cross-modal interactions, we measure how vision enhances or distorts textual grounding by comparing model representations when given structured vision inputs versus degraded or nullified visual signals. To aggregate these effects at a higher level, we perform PCA to extract the primary direction of modality influence across multiple samples. Finally, we develop a dynamic test-time intervention module to adjust the model’s reliance on each modality, effectively reducing hallucination while preserving overall task performance.

Our key contributions are as follows:
\begin{itemize}
\item \textbf{Causal analysis of hallucination.} We identify unintended direct modality influences as a primary cause of hallucination in VLMs.
\item \textbf{Test-time hallucination reduction.} Our method systematically mitigates hallucination by ensuring proper multi-modal fusion and reasoning.
\item \textbf{Superior benchmark performance.} Our approach consistently outperforms existing methods on two VLMs across two diverse benchmarks.
\end{itemize}

\section{Related Works}

\noindent \textbf{Hallucination in Vision-Language Models.}
Recent work has developed VLMs by integrating visual encoders with pre-trained LLMs \cite{instructblip,liu2023visual,zhu2023minigpt}. This allows LLMs to interpret vision tokens from a pre-trained backbone, achieving strong multimodal understanding \cite{NEURIPS2023_407106f4}. However, these models also inherit the LLMs' tendency to generate ungrounded content, commonly termed ``hallucination" \cite{bang2023multitask,huang2021factual,favero2024multi}. 
A major issue in VLM hallucinations is the incorrect inclusion of objects absent from the visual input \cite{bang2023multitask,huang2021factual,li2023evaluating,wang2023evaluation}. Studies suggest this often involves common or co-occurring objects in training data \cite{li2023blip}. Moreover, VLMs struggle with instructions requiring the recognition of absent objects, prompting research on improving model robustness \cite{liu2023aligning}. Some studies attribute hallucinations to object co-occurrence, model uncertainty, and spatial positioning in text, proposing post-hoc correction methods \cite{zhou2023analyzing}.
Hallucination, originally studied in NLP, has become a concern in multimodal models due to its impact on performance \cite{ji2023survey}. Common mitigation strategies rely on additional training to improve alignment with ground truth \cite{yue2024less}, but these methods demand significant data and computation. Training-free alternatives, such as self-feedback correction, auxiliary knowledge models, and enhanced decoding, offer practical solutions but often primarily focus on text rather than addressing vision-induced hallucinations \cite{yin2024woodpecker}.

\noindent \textbf{Causality-Inspired Vision-Language Models.}
Causal inference provides a powerful framework for understanding and controlling the underlying mechanisms in machine learning models. By estimating causal effects, it enables the removal of spurious correlations, disentanglement of meaningful model behaviors, and identification of invariant features that enhance generalization across diverse scenarios \cite{li2022invariant}. Recently, causal methods have been increasingly applied to computer vision, benefiting tasks such as visual explanation \cite{wang2020scout}, image and video recognition \cite{Li2023mm}, scene graph generation \cite{liicassp}, and representation learning \cite{Li_Ji_Wu_Li_Qin_Wei_Zimmermann_2024}. In the context of VLMs, causal analysis is particularly valuable for addressing hallucination, as it allows us to separate genuine multi-modal reasoning from biased modality dominance. By leveraging causal graphs and counterfactual reasoning, we can systematically diagnose and mitigate modality-specific artifacts, ensuring that model predictions are grounded in meaningful cross-modal interactions rather than unintended shortcuts.
\section{Preliminaries}

\begin{figure*}[t]
    \centering
    \begin{subcaptionbox}{Causal graph for traditional single-modal model.\label{fig:single}}[0.3\linewidth]
        {\includegraphics[width=\linewidth]{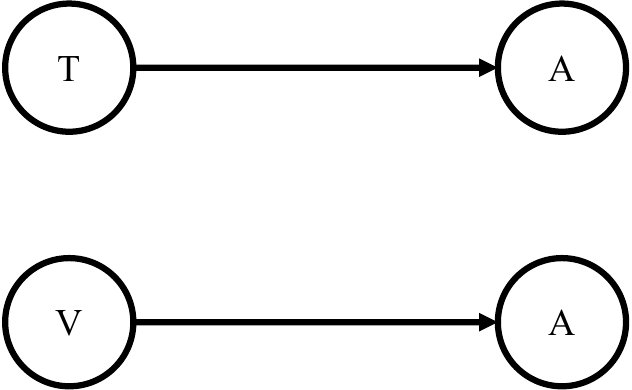}}
    \end{subcaptionbox}
    \hfill
    \begin{subcaptionbox}{Causal graph for Vision-Language Model.\label{fig:vlm}}[0.3\linewidth]
        {\includegraphics[width=\linewidth]{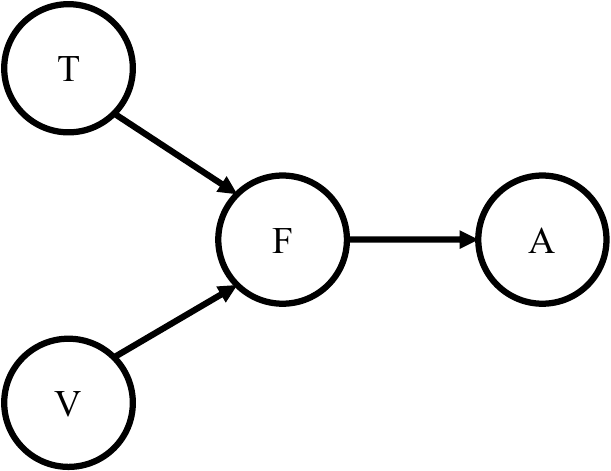}}
    \end{subcaptionbox}
    \hfill
    \begin{subcaptionbox}{Causal graph for biased Vision-Language Model.\label{fig:biased-vlm}}[0.3\linewidth]
        {\includegraphics[width=\linewidth]{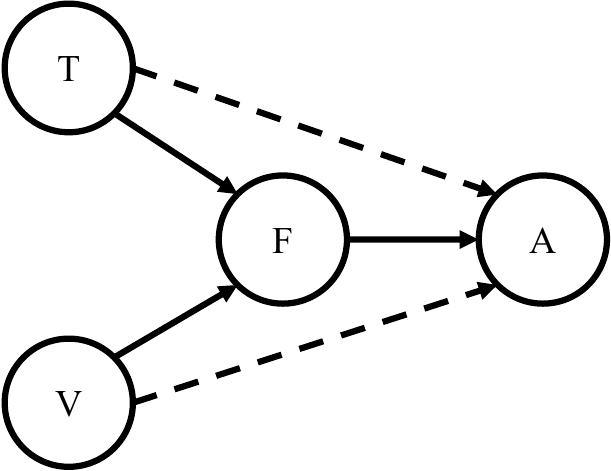}}
    \end{subcaptionbox}
    \caption{Causal graphs for single-modal models and Vision-Language Models (VLMs) are shown. An optimal VLM generates answers conditioned on both vision and text input pairs. However, vision and text inputs may individually exert a direct influence on the output. This direct influence can lead to the hallucination problem in VLMs, where the generated answers are inconsistent with the provided visual or textual context. T: Text input. V: Vision input. A: Answer.}
    \label{fig:SCG}
\end{figure*}

In this section, we propose a series of structural causal graphs (SCGs) (\S \ref{subsec:scg}) for different scenarios to illustrate the superficial correlations between visual inputs, language inputs, and generated answers (\S \ref{subsec:potential_biased}). We then analyze the hallucination problem in VLMs and provide a causal interpretation to explain its underlying causes (\S \ref{subsec:causal_pers}).

\subsection{Structural Causal Graph}
\label{subsec:scg}
The SCGs for different scenarios are illustrated in Fig.~\ref{fig:SCG}. The effects of visual input $V$ and textual input $T$ on the output $A$ can be categorized into two types: single-modal impact (Traditional computer vision tasks or Large Language Models) and multi-modal impact (Vision-Language Models). As shown in Fig.~\ref{fig:single}, the single-modal impact captures the direct influence of $V$ or $T$ on $A$ through $V \rightarrow A$ or $T \rightarrow A$. In contrast, the multi-modal impact represents the indirect effect of $V$ and $T$ on $A$ via the multi-modal fused knowledge $F$, formulated as $(V, T) \rightarrow F \rightarrow A$, as shown in Fig.~\ref{fig:vlm}. The underlying rationale behind the SCG is explained as follows:

\noindent •	$T \rightarrow A$: This represents the data flow in traditional Large Language Models (LLMs), where natural language inputs (typically comprising instructions and data) are processed by the LLM to generate the corresponding output $A$.
    
\noindent •	$V \rightarrow A$: This corresponds to traditional computer vision tasks, such as image captioning, where images are provided as input, and the output $A$ is generated solely based on visual information without language-based context.
    
\noindent •	$(V, T) \rightarrow F \rightarrow A$: This illustrates the mechanism of modern Vision-Language Models. The visual input $V$ is first processed by a vision backbone (e.g., a convolutional neural network or a transformer-based vision encoder) to extract high-level visual features. These visual features are then projected into a shared embedding space compatible with the LLM. Simultaneously, the textual input $T$ is encoded by the LLM. The multi-modal fusion module combines the visual and textual representations to form the fused knowledge $F$. Finally, the LLM leverages this fused knowledge $F$ to generate the answer $A$, integrating both vision and language modalities for coherent and context-aware outputs.

\subsection{Potential Biased Independent Influence}
\label{subsec:potential_biased}
Although the optimal Vision-Language Model is expected to generate answers solely based on the combined vision and text input pairs, in practice, vision and text inputs may still exert direct and independent influences on the output $A$. As illustrated in Fig.~\ref{fig:biased-vlm}, these unintended direct influences are highlighted by dashed arrows, indicating potential shortcut paths that bypass the multi-modal fusion process. Such direct influences can lead to the hallucination problem, where the generated answer $A$ does not align with the provided visual context or textual input.

\noindent •	$T \dashedrightarrow A$: The textual input $T$ may directly influence the output $A$ without considering the visual information. For instance, the model might rely heavily on language priors or contextual cues from the text alone, resulting in answers that ignore relevant visual details. This direct influence can lead to hallucinated responses that appear semantically plausible based on the text but remain inconsistent with the actual visual content.

\noindent •	$V \dashedrightarrow A$: Similarly, the visual input $V$ may directly affect the output $A$ without proper alignment with the textual input. In this scenario, the model might over-rely on visual patterns or features, producing answers that are disconnected from the given textual instructions or questions. This form of direct influence also contributes to hallucinations, where the output appears visually grounded but fails to reflect the intended textual semantics.

These dashed causal paths emphasize the inherent challenge in VLMs: ensuring that the answer $A$ is truly conditioned on the coherent fusion of both $V$ and $T$, rather than being dominated by a single modality. Addressing these unintended direct influences is essential for mitigating hallucination problems and improving the overall reliability and consistency of VLMs.

\subsection{Causal Perspective on VLM Hallucination}
\label{subsec:causal_pers}
From a causal perspective, the hallucination problem in VLMs arises when the model over-relies on a single modality, leading to outputs that are misaligned with the intended multi-modal context. Specifically, unintended direct influences from either the vision or text modality, or their interaction, can dominate the output generation process, causing hallucinated responses. To systematically examine and mitigate these biases, we focus on the \textit{Natural Direct Effect (NDE)} as a means to quantify the direct contributions of each modality and their interaction.

\begin{definition}[Causal Notations] 
Causal notations are used to translate causal assumptions from structural causal graphs into formal mathematical expressions, allowing precise quantification of modality influences on model outputs.
\end{definition}

\begin{definition}[Natural Direct Effects (NDE)] 
The \textit{Natural Direct Effect (NDE)} measures the direct impact of a modality on the output $A$ while holding the multi-modal fusion process consistent. We consider three types of NDEs to capture both the individual and interactive effects of the vision and text modalities:
\end{definition}

Formally, given the causal graph illustrated in Fig.~\ref{fig:biased-vlm}, the answer $A$ is influenced by three paths: $T \rightarrow A$, $V \rightarrow A$, and $F \rightarrow A$. The corresponding causal notation is:

\begin{equation}
    Y_{T, V} = Y(t, v, F(t, v)),
\end{equation}
where $F(\cdot)$ denotes the multi-modal fusion process.

\noindent \textbf{1) Vision Direct Effect (NDE\textsubscript{V}):}  
The direct influence of the vision modality is assessed by altering the vision input while keeping the textual input fixed. Formally:

\begin{equation}
    \text{NDE}(V) = Y(t, v, F(t, v)) - Y(t, v_{*}, F(t, v_{*})),
\end{equation}
where $v$ denotes the original vision input and $v_{*}$ represents the treated vision input. This formulation captures how much the vision modality alone contributes to the output, independent of multi-modal fusion consistency.

\noindent \textbf{2) Text Direct Effect (NDE\textsubscript{T}):}  
The direct influence of the text modality is measured by modifying the textual input while keeping the visual input constant:

\begin{equation}
    \text{NDE}(T) = Y(t, v, F(t, v)) - Y(t_{*}, v, F(t_{*}, v)),
\end{equation}
where $t$ is the original text input and $t_{*}$ represents the treated text input. This equation reflects how text alone influences the output, independent of visual grounding.

\noindent \textbf{3) Cross-Modality Direct Effect (NDE\textsubscript{V,T}):}  
While the vision modality treatment assesses the direct influence of vision by altering visual inputs, it does not capture how vision complements textual information in multi-modal reasoning. In practice, vision often provides contextual cues that enhance text interpretation. Thus, it is essential to evaluate how vision interacts with text to influence the output.

To this end, we propose the \textit{Cross-Modality Direct Effect (NDE\textsubscript{V,T})}, which quantifies the complementary role of vision when combined with text. Unlike vision treatment, which isolates vision’s standalone contribution, this analysis evaluates scenarios where textual input is paired with a partially informative image versus a non-informative one. The formulation is:

\begin{equation}
    \text{NDE}(V, T) = Y(t, v_{*}, F(t, v_{*})) - Y(t, v_{\text{null}}, F(t, v_{\text{null}})),
\end{equation}
where $v_{\text{null}}$ denotes a non-informative visual input. A high $\text{NDE}(V, T)$ indicates meaningful visual-textual complementarity, while a low or negative value suggests that vision introduces noise, potentially leading to hallucinations.

By focusing on these direct effects, our causal analysis framework provides a clear diagnostic approach to understanding and mitigating hallucination in VLMs. This framework highlights the necessity of balanced multi-modal fusion, where each modality contributes appropriately to the final prediction without dominating the reasoning process.
\section{Methodology}
\renewcommand{\arraystretch}{1.5}
\begin{table*}[t]
    \centering
    \vspace{5pt}

    \begin{tabular}{c c c c c c c c c c}  
        \toprule
        \multirow{2}{*}{\textbf{Settings}} & \multirow{2}{*}{\textbf{Method}} & \multicolumn{4}{c}{\textbf{LLaVA 1.5}} & \multicolumn{4}{c}{\textbf{InstructBlip}} \\
        \cmidrule(lr){3-6} \cmidrule(lr){7-10}
        & & Accuracy & Precision & Recall & F1 score & Accuracy & Precision & Recall & F1 score \\
        \hline
        \multirow{4}{*}{Random} & Regular &83.49  &88.83  &76.70  &82.34  &80.42  &78.93  &83.21  &81.01  \\ \cline{2-10}
        & VCD &86.84  &87.15  &\underline{86.68}  &\underline{86.91}  &84.10  &84.21  &\underline{85.36}  &\underline{84.78}  \\ \cline{2-10}
        & Opera &\underline{87.53}  &\textbf{94.52}  &79.80  &86.53  &\underline{85.07}  &\textbf{88.39}  &80.73  &84.39  \\ \cline{2-10}

        & Our Method & \textbf{89.10}  &\underline{90.59}  &\textbf{87.27}  &\textbf{88.89}  &\textbf{88.83}  &\underline{88.04}  &\textbf{89.87}  &\textbf{88.95}  \\ 
        \hline
        \multirow{4}{*}{Popular} & Regular &79.98  &82.47  &76.72  &79.48  &76.10  &73.22  &82.94  &77.78  \\ \cline{2-10}
        & VCD &82.65  &87.15  &\underline{80.60}  &\underline{83.74}  &\underline{79.94}  &\underline{77.84}  &83.33  &\underline{80.49}  \\ \cline{2-10}
        & Opera &\underline{84.21}  &\textbf{88.00}  &79.80  &83.70  &78.33  &73.85  &\underline{87.73}  &80.19  \\ \cline{2-10}

        & Our Method &\textbf{87.53}  &\underline{87.73}  &\textbf{87.27}  &\textbf{87.50}  &\textbf{83.27}  &\textbf{79.39}  &\textbf{89.87}  &\textbf{84.30}  \\ 
        \hline
        \multirow{4}{*}{Adversarial} & Regular &76.03  &76.11  &76.80  &76.45  &72.37  &68.78  &83.06  &75.24  \\ \cline{2-10}
        & VCD &77.31  &73.43  &\underline{86.47}  &79.42  &\textbf{76.32}  &\textbf{73.24}  &84.08  &\underline{78.29}  \\ \cline{2-10}
        & Opera &\underline{80.88}  &\textbf{82.16}  &79.76  &\underline{80.94}  &75.50  &70.50  &\underline{87.73}  &78.17  \\ \cline{2-10}
  
        & Our Method &\textbf{81.70}  &\underline{78.90}  & \textbf{87.27} &\textbf{82.87}  &\underline{76.23}  &\underline{70.84}  &\textbf{89.87}  &\textbf{79.22}  \\ 
        \bottomrule
    \end{tabular}

    \caption{Performance comparison on POPE (Regular, Popular, and Adversarial) across two state-of-the-art Vision-Language Models (LLaVA 1.5 and InstructBlip). The best performance in each column is indicated in bold, and the second-best is underlined. Our proposed causal intervention method consistently outperforms existing methods (VCD, Opera), demonstrating improved accuracy and reduced hallucination across different evaluation settings.}
    \label{tab:performance}
\end{table*}

Building on prior work in editing vision-language model intermediate representations \cite{liu2024reducing,jiang2024interpreting}, we quantify the \textit{Natural Direct Effects (NDEs)} of different modalities by analyzing representation shifts before and after applying modality-specific perturbations. This allows us to analyze separately the contributions of vision and text, along with their interaction, to the final model output.

\noindent \textbf{Measuring NDE\textsubscript{V}.}
To measure the vision modality’s direct effect, we introduce perturbations to the visual input and assess their impact on representations. 

Given an image input $I$, we extract its vision representation $V_{i,k}^{I}$ from the $i$-th layer at the $k$-th visual token. We then apply $m$ different random masks, $C_j$ for $j \in \{1, \dots, m\}$, to corrupt the image, producing masked versions $M_j(I)$. The vision encoder processes each perturbed input $M_j(I)$, yielding the corresponding representations $V_{i,k}^{M_j(I)}$. To estimate the perturbed vision representation, we take the average of these masked representations as $\bar{V}_{i,k}^{I}$.

The direct effect of the vision modality for the image $I$ is then quantified as the difference between the original and perturbed representations:

\begin{equation}
    D_{i,k}^{I} = \bar{V}_{i,k}^{I} - V_{i,k}^{I}.
\end{equation}

To obtain a global-level estimate of $\text{NDE}_V$ (as opposed to the instance-level effect $D_{i,k}^{I}$), we sample $N$ images and compute their respective direct effects, systematically stacking them into a structured matrix:

\begin{equation}
    [D_{i,k}^{I_{1}}, D_{i,k}^{I_{2}}, ..., D_{i,k}^{I_{N}}].
\end{equation}

Following \citet{liu2024reducing}, we perform PCA on this matrix and use the first principal direction as the global-level estimate of $\text{NDE}_V$.

\noindent \textbf{Measuring NDE\textsubscript{T}.}
To measure the direct effect of the text modality, we introduce controlled textual hallucinations and analyze their influence on representations.  

We randomly sample $N$ image captions $C_N$ and generate their hallucinated counterparts $C_N^{h}$ using a GPT model. For each caption, we extract the last-token representation from the $i$-th layer, denoted as $T_{i}^{C_{N}}$ for the original text and $T_{i}^{C_{N}^{h}}$ for the hallucinated version. The  direct effect of text modality can be computed as:

\begin{equation}
    D_{i}^{T} = T_{i}^{C_{N}^{h}} - T_{i}^{C_{N}}.
\end{equation}

To estimate global-level $\text{NDE}_T$, we stack the text direct effect vectors for all sampled captions into a matrix and apply PCA, obtaining the first principal direction as the final measure of $\text{NDE}_T$.

\noindent \textbf{Measuring NDE\textsubscript{V,T}.}
To quantify the cross-modality direct effect of vision and text, we evaluate how vision complements textual information in multi-modal reasoning. Unlike $NDE_{V}$, which isolates vision’s standalone impact, $NDE_{V,T}$ comprehensively captures the extent to which vision enhances or distorts textual semantic grounding.

We begin by sampling $N$ images $I_N$ and their corresponding textual descriptions $C_N$. For each image, we generate two perturbed versions:  
1) $I_{\text{black}}$ — a fully black image, containing no meaningful visual information. This setting ensures that the vision encoder receives an input with no structured content while preserving input dimensions and format.  
2) $I_{\text{null}}$ — a no-input condition, where the model receives no visual input at all. This serves as an extreme reference case to thoroughly assess the model’s reliance on textual information alone.

For each case, we obtain the visual representations $V_{i,k}^{I_{\text{black}}}$ and $V_{i,k}^{I_{\text{null}}}$ at the $i$-th layer and $k$-th token. The cross-modality direct effect is then defined as:

\begin{equation}
    D_{i,k}^{V,T} = V_{i,k}^{I_{\text{black}}} - V_{i,k}^{I_{\text{null}}}.
\end{equation}

A high $\text{NDE}_{V,T}$ suggests that vision provides complementary information to text, improving multi-modal understanding. Conversely, a low or negative $\text{NDE}_{V,T}$ suggests that vision introduces noise or misalignment, potentially leading to hallucinated responses.

For global-level analysis, we stack the cross-modality direct effect vectors across $N$ samples and apply PCA, using the first principal direction as the final estimate of $\text{NDE}_{V,T}$.  

\noindent \textbf{Test-time Intervention.}  
We integrate the computed Natural Direct Effects, $NDE_V$, $NDE_T$, and the cross-modal component $NDE_{V,T}$, to adjust the outputs of both the vision and text encoders during inference. Specifically, we modify the intermediate representations at every layer and token position as follows:
\begin{align}
    V_{i,k}^{I'} &= V_{i,k}^{I} + a \cdot NDE_V, \\
    T_{i}^{C_{N}'} &= T_{i}^{C_{N}} + b \cdot NDE_{V,T} + c \cdot NDE_T.
\end{align}

This test-time intervention approach is highly efficient and can be seamlessly integrated into all major VLM frameworks and architectures.
\section{Experiments}
\renewcommand{\arraystretch}{1.2}
\begin{table*}[t]
    \centering
    \vspace{5pt}
    \resizebox{1\textwidth}{!}{
    \begin{tabular}{c c c c c c c c c c}  
        \toprule
        \textbf{Method} & \textbf{Average} & Attribute & Adversarial & Comparison & Counting & Relation & Environment & Holistic & Other \\
        \midrule
        Regular &2.06 &\underline{3.25} &1.83 &2.25 &2.40 &1.83  &1.92 &1.67 &1.33        \\ \cline{1-10}
        VCD &\underline{2.69}  & \underline{3.25} & \underline{2.18}  &\underline{3.00}  &\textbf{2.42} &\underline{2.58} & \underline{3.25} & \textbf{2.42} &\underline{2.42}   \\ \cline{1-10}
        Opera &2.64  &2.92& \textbf{2.25} &  2.75 &\underline{2.41} &\textbf{2.92} &\textbf{3.26} & \underline{2.33} & 2.25     \\ \cline{1-10}
        
        Our Method &\textbf{2.82} &\textbf{4.00} &2.17& \textbf{3.83} & 2.25 & 2.42 & 2.83 & \textbf{2.42} & \textbf{2.67}\\  \cline{1-10}
        \bottomrule
    \end{tabular}
    }
    \caption{Performance comparison on MMHal-Bench with LLaVA 1.5. The best performance in each column is indicated in bold, and the second-best is underlined. Our proposed causal intervention method consistently outperforms existing methods (VCD, Opera), demonstrating improved accuracy and reduced hallucination across different evaluation settings.}
    \label{tab:mmhal_performance}

\end{table*}

\subsection{Datasets and Evaluation Metrics}
\label{subsec:dataset_evaluation}

\noindent \textbf{Datasets.}
We evaluate on two visual hallucination task benchmarks: (1) MMHal-Bench~\cite{sun2023aligning}, and (2) POPE~\cite{li2023evaluating}. See details in Appx.~\ref{appx:addi_set}.

\noindent \textbf{Evaluation Metrics.}
We conduct evaluations following the original evaluation method.
(1) \textbf{MMHal-Bench:} According to the evaluation results in MMHal-Bench, GPT-4~\cite{openai2023gpt4} can achieve a 94\% agreement rate with human judgments. Therefore, we use GPT-4o-mini~\cite{openai2024gpt4ocard} to analyze and score the responses of LMMs.  
Following the assessment method in MMHal-Bench, we provide GPT-4o-mini with the question and the VLM's response. Additionally, we supply the category name of the image content and a standard human-generated answer to improve the accuracy of response evaluation. Ultimately, GPT-4o-mini returns the VLM's scores across the 8 question categories and its hallucination rate.
(2) \textbf{POPE:} Since POPE consists entirely of Yes/No questions, the correctness of VLM responses can be directly determined based on the ground-truth answers. This allows for the calculation of accuracy, precision, recall, and F1 score, with F1 score serving as the primary metric.

\begin{figure*}[h]
\centering
\includegraphics[width=0.5\linewidth]{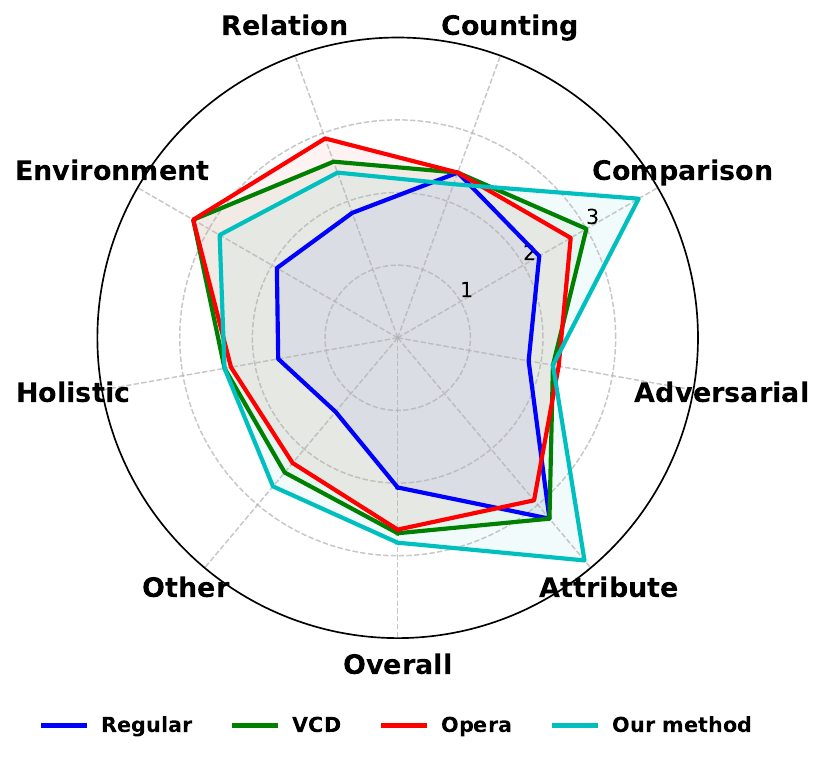}
\caption{Overall performance and detailed score of different methods on the 8 question categories of MMHal-Bench. Our method achieves the best overall performance and significantly outperforms existing methods (VCD, Opera) in Attribute and Comparison.}
\label{fig:mmhal_radar}
\end{figure*}

\subsection{Implementation Details}
We evaluate the effectiveness of our method on two widely used 7B VLMs, LLaVA 1.5~\cite{liu2023visual} and InstructBLIP~\cite{instructblip}. Additionally, we evaluate our method against two state-of-the-art baselines for alleviating hallucinations in the decoding stage: VCD~\cite{Leng_2024_CVPR} and Opera~\cite{huang2024opera}. Our default hyperparameter is sampling size \( N = 50 \). To ensure a fair comparison, we set $a = b = c = 0.9$ for both models across all experiments.
 All experiments are conducted using PyTorch with Nvidia RTX A6000 GPUs.

\renewcommand{\arraystretch}{1.2}
\begin{table*}[t]
    \centering
    \vspace{5pt}
    \resizebox{0.9\textwidth}{!}{
    \begin{tabular}{c c c c c c c c c c}  
        \toprule
        \textbf{PCA dim} & \textbf{Average} & Attribute & Adversarial & Comparison & Counting & Relation & Environment & Holistic & Other \\
        \midrule
        Regular &2.06 &3.25 &1.83 &2.25 &2.40 &1.83  &1.92 &1.67 &1.33        \\ \cline{1-10}
        1 &2.82 &4.00 &2.17 &3.83 &2.25 &2.42 &2.83 &2.42 &2.67        \\ \cline{1-10}
        3 &2.51 &3.58&1.67&3.58&1.92&2.5&3.08&1.67&2.08   \\ \cline{1-10}
        5 &2.42 &3.58&1.67&3.08&1.75&2.08&3.08&1.58&2.5     \\ \cline{1-10} 
        \bottomrule
    \end{tabular}
    }
    \caption{Performance of LLaVA 1.5 on MMHal-Bench with different PCA dimensions. `Regular' denotes the baseline method without any enhancement.}
    \label{tab:PCA_performance}
\end{table*}
\renewcommand{\arraystretch}{1.2}
\begin{table}[t]
    \centering
    \vspace{5pt}
    \resizebox{0.5\textwidth}{!}{
    \begin{tabular}{c |c c}  
        \toprule
        \textbf{Number of samples} & \textbf{Average$\uparrow$} & \textbf{Hallucination rate$\downarrow$}\\
        \midrule
        Regular  & 2.06 & 64.58 \\ \cline{1-3}
        25  & 2.45 & 51.04 \\ \cline{1-3}
        50  & 2.82 & 45.83 \\ \cline{1-3}
        75  & 2.62 & 45.83\\ \cline{1-3}
        100 & 2.58 & 50.00\\ 
        \bottomrule
    \end{tabular}
    }
    \caption{Performance of LLaVA 1.5 on MMHal-Bench with different numbers of samples. `Regular' denotes the baseline method without any enhancement.}
    \label{tab:Sample_performance_small}

\end{table}

\subsection{Experimental Results}
\label{subsec:experiment}

Tab.\ref{tab:performance}, Tab.\ref{tab:mmhal_performance}, and Fig.~\ref{fig:mmhal_radar} demonstrate the effectiveness of our method compared to the State-of-the-Art approaches in two VLMs and benchmarks. Our method consistently achieves top or near-top results in all metrics. See more analysis in Appx.~\ref{appx:addi_ex}.

Results from Tab.~\ref{tab:performance} highlight key trends across Random, Popular, and Adversarial settings for LLaVA 1.5 and InstructBlip. In the Random setting, our method significantly improves accuracy (e.g., 83.49 to 89.10 in LLaVA 1.5) and recall (76.70 to 87.27), demonstrating the effectiveness of removing unintended direct modality influences. In the Popular setting, our method mitigates reliance on language priors, leading to higher accuracy (e.g., 79.98 to 87.53 in LLaVA 1.5) and F1 scores. Under the challenging Adversarial setting, our approach remains robust, significantly improving recall (76.80 to 87.27 in LLaVA 1.5) and F1 scores. These results validate that our causal intervention mechanism systematically reduces hallucination while enhancing resilience in diverse conditions.

Tab.~\ref{tab:mmhal_performance} further demonstrates our method’s superiority across MMHal-Bench categories, achieving the highest average performance (2.82) over VCD (2.69) and Opera (2.64). It excels in Attribute (4.00), Comparison (3.83), and Other (2.67) categories, indicating enhanced multi-modal reasoning. While slightly behind in Adversarial and Counting, our method remains competitive. Strong performance in Holistic (2.42) and Environment (2.83) categories confirms that reducing unintended modality influences improves vision-text alignment.

Overall, our causal intervention framework effectively reduces hallucination, leading to more accurate and reliable multi-modal reasoning across diverse tasks. These results underscore the importance of addressing unintended modality biases in VLMs to improve robustness.

\subsection{In-Depth Analysis}

\noindent \textbf{Measuring NDE with Different PCA Dimensions.}
Tab.~\ref{tab:PCA_performance} shows that using a single principal component (PCA dim = 1) yields the highest overall performance (2.82), outperforming PCA dim = 3 (2.51) and PCA dim = 5 (2.42). This suggests that restricting modality influence to a single direction effectively mitigates hallucinations while preserving multi-modal reasoning. Performance declines in Adversarial (from 2.17 to 1.67) and Holistic (2.42 → 1.58) categories with higher PCA dimensions indicate that excessive components may reintroduce noise, weakening robustness and interpretability. These results highlight that a minimal but targeted reduction in the influence of the modality enhances the accuracy of the reasoning.

\noindent \textbf{Effect of Sample Size.}
As shown in Tab.~\ref{tab:Sample_performance_small}, using 50 samples achieves the best performance (2.82), outperforming both smaller (25 samples, 2.45) and larger settings (75 and 100 samples). Gains are most evident in Attribute (4.00) and Comparison (3.83), indicating improved hallucination mitigation. Performance drops at 75 and 100 samples suggest redundancy or overfitting, particularly in Adversarial and Holistic categories. These findings indicate that an optimal sample size (50) ensures robust estimation of modality influences while avoiding excessive noise, leading to better reasoning and reduced hallucinations.

\noindent \textbf{Qualitative Analysis.}
\label{subsec:qualitative_ana}
To further demonstrate the effectiveness of our approach, we provide extensive visualizations comparing outputs before and after applying our method. These qualitative examples highlight reductions in hallucination and improved alignment with visual context. Detailed case studies can be found in the appx.~\ref{appx.qualitative}.
\section{Conclusion, Limitations, and Future Work}
In this work, we introduced a causal framework to analyze and mitigate hallucination in VLMs. By constructing structural causal graphs and estimating the Natural Direct Effect of each modality, we identified unintended direct modality influences as a key contributor to hallucination. Our proposed test-time intervention mechanism effectively reduces modality bias, ensuring that generated outputs are more accurately grounded in fused multi-modal information. Empirical results across multiple benchmarks demonstrate that our method improves the reliability of VLMs while maintaining task performance.

\noindent \textbf{Limitations and Future Works.} 
The causal framework may not capture all hallucination sources, especially in open-ended tasks. Additionally, the intervention introduces inference overhead, impacting real-time use. Future work can refine the causal model, develop task-specific adaptive interventions, and integrate contrastive learning for better multi-modal alignment.

\noindent \textbf{Broader Impact and Ethics Statement.}

\noindent \textbf{Broader Impact Statement:}
Our method improves the reliability of the VLM by reducing hallucinations and improving trust in AI applications such as healthcare and autonomous systems. However, it does not eliminate biases in training data, and strict hallucination control may limit creative applications. Future work should balance factual consistency with flexibility across different use cases.

\noindent \textbf{Ethics Statement:}
This research improves the factual grounding of VLM without altering training data. Although our approach reduces hallucination, it does not guarantee complete accuracy, requiring users to apply additional validation in sensitive applications. Responsible deployment is essential to effectively prevent misuse or excessive overreliance on AI-generated outputs.

{
    \small
    \bibliographystyle{ieeenat_fullname}
    \bibliography{main}
}

\appendix
\newpage

\setcounter{table}{0}
\setcounter{figure}{0}

\renewcommand{\thetable}{\Alph{table}}
\renewcommand{\thefigure}{\Alph{figure}}

\section{Additional Experimental Settings}
\label{appx:addi_set}
As briefly discussed in \S \ref{subsec:dataset_evaluation}, we evaluate our method on two benchmarks. 

(1) \textbf{MMHal-Bench}~\cite{sun2023aligning} is designed to evaluate hallucinations in VLMs' responses. It includes 96 image-question pairs across 8 question categories and 12 object topics from MSCOCO~\cite{lin2014microsoft}. It specifically targets types of questions where VLMs are prone to making false claims about image content, including object attributes, adversarial objects, comparison, counting, spatial relations, environment, holistic description, and other cases, such as misreading text or icons. Evaluation is conducted using GPT-4o-mini, which compares model responses against human-generated answers to determine hallucination presence, and additional context is provided to enhance its judgment. 

(2) \textbf{POPE}~\cite{li2023evaluating} (Polling-based Object Probing Evaluation) is a polling-based evaluation benchmark for assessing object hallucination in VLMs. It formulates the evaluation of object hallucination as a binary classification task by prompting VLMs with questions that require “Yes” or “No” responses. POPE maintains a balanced distribution, ensuring an equal split between queries for existing and non-existing objects, and utilizes three sampling strategies: random, popular, and adversarial. It collects 500 images from each of the MSCOCO ~\cite{lin2014microsoft}, A-OKVQA ~\cite{schwenk2022okvqa}, and GQA ~\cite{hudson2019gqa}, and then samples objects that VLMs are prone to hallucinate, generating a total of 27,000 challenging Yes/No questions to assess the model’s ability to correctly identify objects in images. POPE adopts Accuracy, Precision, Recall, and F1-score as evaluation metrics.

\begin{figure*}[t]
    \centering
    \includegraphics[width=\textwidth]{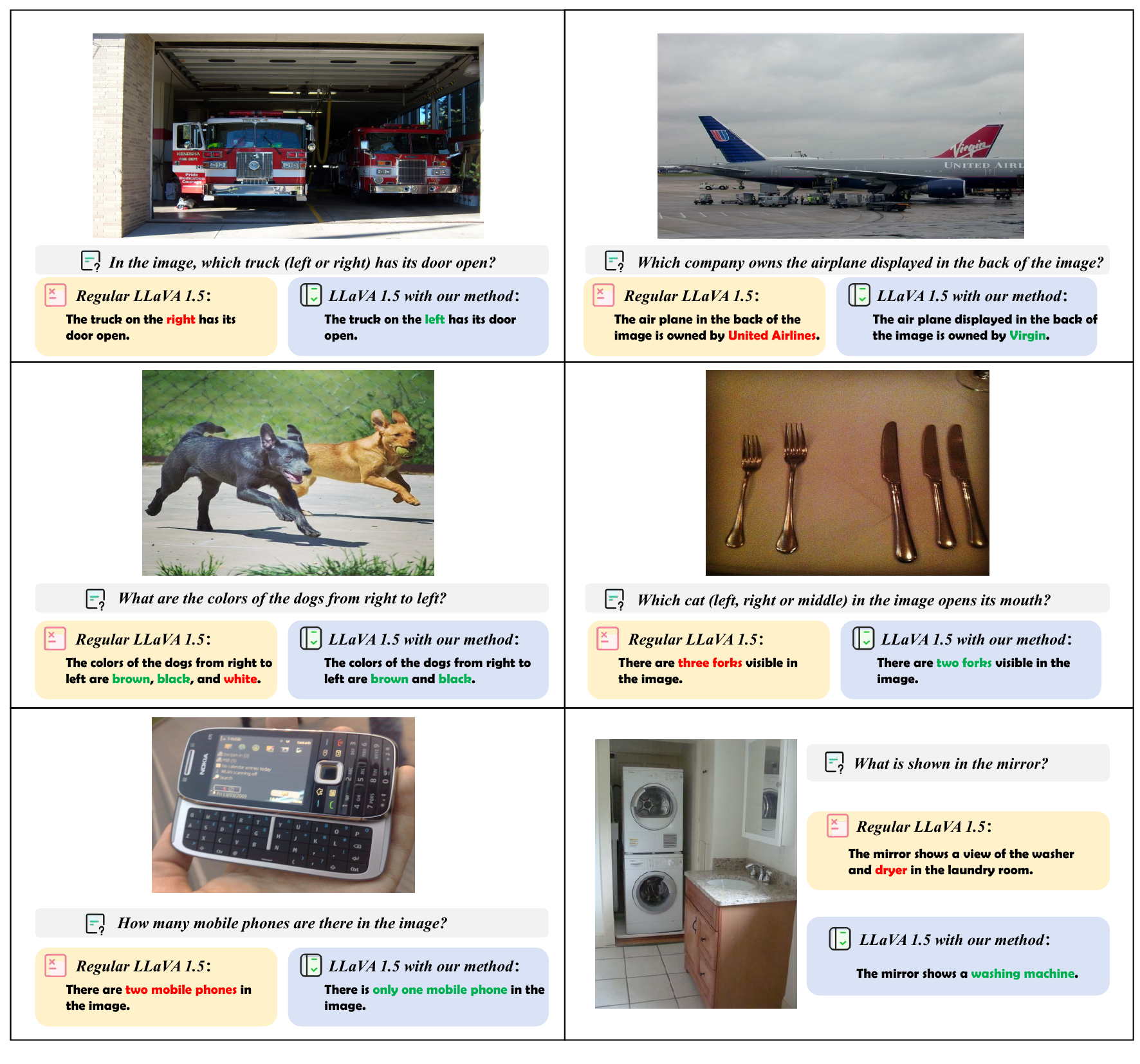}
    \caption{Case study illustrating the impact of our method on VLM hallucination. The figure compares outputs from the original model and our enhanced approach, highlighting reductions in hallucinated content and improved alignment with the visual context. Our method effectively mitigates incorrect textual descriptions by refining modality interactions, leading to more accurate and reliable multi-modal reasoning.}
    \label{appx:fig_case_study}
\end{figure*}

\section{Additional Experimental Analysis}
\label{appx:addi_ex}

As briefly discussed in \S \ref{subsec:experiment}, we evaluate our method on two benchmarks. 

The results summarized in Tab.~\ref{tab:performance} reveal several notable trends when comparing our proposed method to existing approaches across Random, Popular, and Adversarial settings for both LLaVA 1.5 and InstructBlip.
Under the Random setting, our method achieves a clear advantage. For instance, with LLaVA 1.5, accuracy increases from 83.49 in the Regular baseline to 89.10, while recall improves from 76.70 to 87.27. In InstructBlip, similar gains are observed: accuracy rises from 80.42 to 88.83, and recall from 83.21 to 89.87. These improvements indicate that our test-time intervention module, which systematically estimates and removes the unintended direct influences from each modality, effectively reduces hallucinations and leads to better alignment between the generated outputs and the intended multi-modal context.
In the Popular setting, our approach again outperforms the alternatives. For LLaVA 1.5, our method boosts accuracy from 79.98 (Regular) to 87.53 and enhances the F1 score from 79.48 to 87.50. InstructBlip also benefits, with accuracy improving from 76.10 to 83.27 and F1 score rising from 77.78 to 84.30. These results suggest that by mitigating the model's over-reliance on language priors and counteracting spurious correlations present in the training data, our method promotes a more balanced integration of visual and textual cues.
The most challenging conditions are observed under the Adversarial setting. Here, the LLaVA 1.5 model's recall jumps significantly from 76.80 to 87.27, and the F1 score improves from 76.45 to 82.87. Although the improvements in InstructBlip are more modest in terms of accuracy (from 72.37 to 76.23), both recall and F1 scores show meaningful enhancements. This pattern indicates that our approach is robust even when the input signals are intentionally degraded or perturbed, highlighting its potential for real-world applications where input quality may vary.
Overall, the experimental data suggest that our causal intervention mechanism—grounded in counterfactual analysis and Natural Direct Effect estimation—is effective in systematically reducing hallucination in VLMs. By eliminating unintended direct modality influences, our method not only improves the accuracy of vision-text fusion but also enhances the model’s resilience across diverse and challenging scenarios.

The experimental results presented in Table~\ref{tab:mmhal_performance} demonstrate the effectiveness of our proposed causal intervention approach in mitigating hallucination and improving the accuracy of vision-language models (VLMs) across multiple reasoning categories in the MMHal-Bench benchmark. Compared to existing methods, our approach consistently achieves the highest average performance score (2.82), outperforming both VCD (2.69) and Opera (2.64), as well as the regular baseline (2.06).
A closer examination of the category-wise results reveals that our method exhibits notable improvements in specific reasoning types. In particular, it achieves the highest performance in Attribute (4.00), Comparison (3.83), and Other (2.67) categories. The superior performance in Attribute reasoning suggests that our method enhances the model’s ability to accurately associate visual details with textual descriptions, a critical factor in reducing hallucinated object properties. Similarly, the strong performance in Comparison tasks indicates improved cross-instance reasoning, likely due to our causal intervention strategy, which ensures that both visual and textual modalities contribute meaningfully to the generated response rather than relying on language priors.
In contrast, while our method does not achieve the highest score in Adversarial, Counting, and Relation categories, it remains competitive, showing marginal differences from the top-performing methods. For instance, in the Adversarial category, our score (2.17) is comparable to Opera (2.25), suggesting that while causal intervention reduces hallucination, certain adversarial perturbations may still challenge the model’s robustness. Additionally, in Counting (2.25), our approach is slightly lower than VCD (2.42), possibly indicating that direct modality influence alone may not fully address numerical inconsistencies, which often require improved object permanence reasoning.
Importantly, our approach demonstrates a balanced improvement across multiple reasoning types, particularly excelling in categories where multi-modal fusion plays a crucial role, such as Holistic (2.42) and Environment (2.83). These results support our hypothesis that hallucination arises due to unintended direct influences from individual modalities, and by systematically mitigating these effects, our method enhances the model’s ability to generate more reliable and contextually grounded outputs.
Overall, these findings validate the effectiveness of our causal intervention framework in reducing hallucination and improving reasoning accuracy across diverse evaluation settings. The performance gains across multiple reasoning categories highlight the necessity of explicitly addressing unintended modality biases in VLMs, reinforcing the potential of causal analysis as a key tool in advancing the robustness of multi-modal models.

\section{Qualitative Result}
\label{appx.qualitative}
As briefly discussed in \S \ref{subsec:qualitative_ana}, we provide more qualitative results to showcase the effectiveness of our method, as shown in Fig.~\ref{appx:fig_case_study}.

\end{document}